\pdfoutput=1

\documentclass[11pt]{article}

\usepackage{ACL2024}

\usepackage{times}
\usepackage{latexsym}

\usepackage[T1]{fontenc}

\usepackage[utf8]{inputenc}

\usepackage{microtype}

\usepackage{inconsolata}
\usepackage{booktabs}
\usepackage{multirow}
\usepackage{float}
\usepackage{adjustbox}
\usepackage{caption}
\usepackage{subcaption}
\usepackage{comment}
\usepackage{siunitx}
\usepackage{amsmath}
\usepackage{mathtools}
\usepackage{tcolorbox}
\usepackage{caption}
\usepackage{varwidth}
\newcommand{\metric}{FENICE}
\newcommand{\dataset}{Story-SummEval}

\newcommand{\argmax}{\mathop{\mathrm{argmax}}}

\title{FENICE: \underline{F}actuality \underline{E}valuation of Summarization Based on \\\underline{N}atural Language \underline{I}nference and \underline{C}laim \underline{E}xtraction}

 \author{Alessandro Scir\`e$^{1,2}$ \qquad Karim Ghonim$^{2}$ \qquad Roberto Navigli$^{2}$\\\\
\qquad Babelscape, Italy  \qquad \qquad ~~~~~
         Sapienza University of Rome \\
  ~$^1$\texttt{scire@babelscape.com} \qquad  
         ~~~$^2$\texttt{\{lastname\}@diag.uniroma1.it}}

\begin{document}
\maketitle
\begin{abstract}
Recent advancements in text summarization, particularly with the advent of Large Language Models (LLMs), have shown remarkable performance. However, a notable challenge persists as a substantial number of automatically-generated summaries exhibit factual inconsistencies, such as hallucinations.
In response to this issue, various approaches for the evaluation of consistency for summarization have emerged. Yet, these newly-introduced metrics face several limitations, including lack of interpretability, focus on short document summaries (e.g., news articles), and computational impracticality, especially for LLM-based metrics.
To address these shortcomings, we propose Factuality Evaluation of summarization based on Natural language Inference and Claim Extraction (\metric), a more interpretable and efficient factuality-oriented metric. \metric~leverages an NLI-based alignment between information in the source document and a set of atomic facts, referred to as \textit{claims}, extracted from the summary. Our metric sets a new state of the art on \textsc{AggreFact}, the de-facto benchmark for factuality evaluation. Moreover, we extend our evaluation to a more challenging setting by conducting a human annotation process of long-form summarization.
In the hope of fostering research in summarization factuality evaluation, we release the code of our metric and our factuality annotations of long-form summarization at \url{https://github.com/Babelscape/FENICE}.
\end{abstract}

\section{Introduction}
In recent years, Natural Language Generation (NLG) approaches have achieved impressive results in many areas of Natural Language Processing (NLP), including Machine Translation (MT), and Question Answering (QA). When it comes to Text Summarization, the landscape differs somewhat. Despite the high performance boasted by conventional metrics such as ROUGE \citep{lin-2004-rouge}, several empirical investigations \citep{maynez-etal-2020-faithfulness, durmus-etal-2020-feqa, kryscinski-etal-2020-evaluating} 
revealed that a significant portion of automatically-generated summaries contains factual inconsistencies, e.g., hallucinations. Addressing this issue is crucial, as it stands as a barrier preventing the widespread application of such systems in real-world scenarios.

With the emergence of Large Language Models (LLMs) the challenge still persists, as their outputs continue to exhibit inconsistencies~\citep{tam2022evaluating}. What is concerning is that the heightened fluency boasted by these models~\cite{wang2023elementaware} may inadvertently lead inexpert users to place excessive trust in their outputs, thereby fostering the spread of misinformation. For these reasons, the introduction of reliable automatic metrics that detect factual inaccuracies in summaries is becoming increasingly urgent.

As a response to this need, various factuality metrics have emerged, leveraging methodologies rooted in Question Answering \citep{scialom-etal-2021-questeval, fabbri-etal-2022-qafacteval, wang-etal-2020-asking}, Natural Language Inference \citep[NLI]{zha-etal-2023-alignscore, chen2023menli, laban-etal-2022-summac}, and semantic representations \citep{ribeiro-etal-2022-factgraph, goyal-durrett-2020-evaluating, fan2023evaluating}. The latest trend in this domain involves works \citep{fu2023gptscore, liu2023geval, gao2023humanlike} that introduce prompt engineering approaches using LLMs.

However, current factuality metrics suffer from one or more of the following drawbacks: i) they often assign a broad score to the entire summary without specifying which parts are factual or hallucinated, thereby lacking interpretability; ii) they are mainly designed for evaluating summaries of short documents, such as news articles, and iii) they are impractical from a computational standpoint, specifically in the case of LLM-based metrics.

\begin{figure*}[t]
    \centering
    \includegraphics[width=1.0\linewidth]{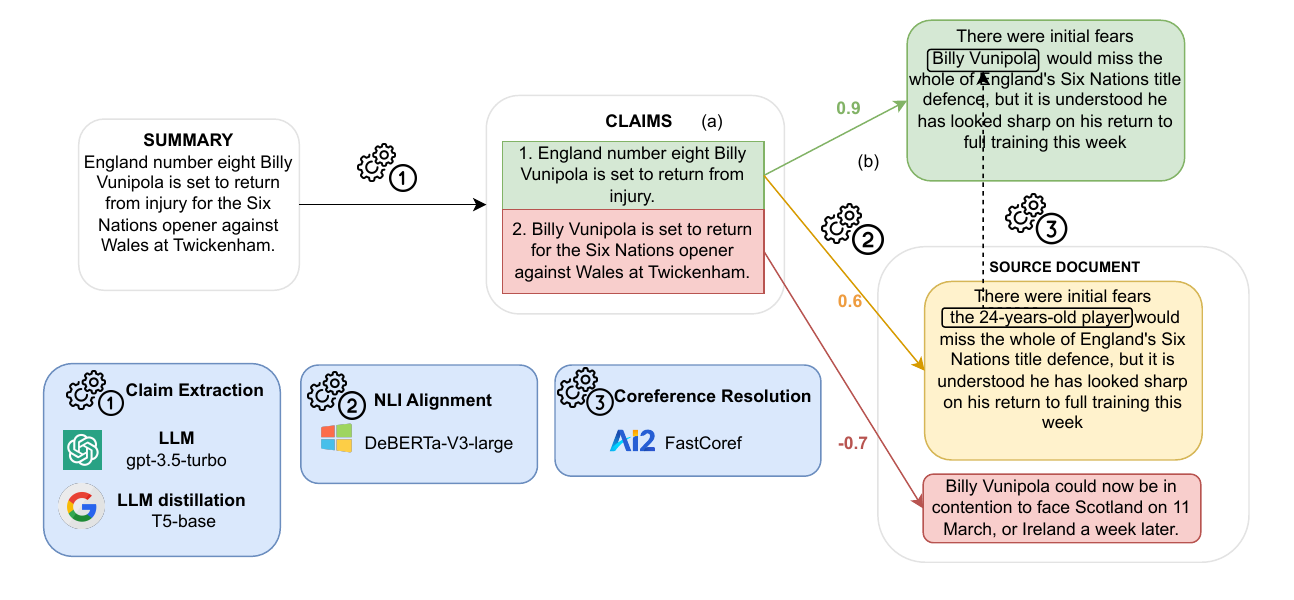}
    \caption{Overview of \metric: the process begins with the extraction of claims from a given summary (step 1). Extracted claims are then aligned with specific sections of the input document (step 2). Finally, we refine the obtained alignments through a coreference-resolution-based approach (step 3). Best seen in color.}
    \vspace{-2mm}
    \label{fig:overview}
\end{figure*}

To overcome these limitations, we introduce Factuality Evaluation of summarization based on Natural language Inference and Claim Extraction (\metric), a state-of-the-art factuality metric that is more interpretable and efficient. As depicted in Figure \ref{fig:overview}, {\metric} evaluates the factuality of a summary by leveraging an NLI-based alignment between information in the source document and a collection of atomic facts, referred to as \textit{claims}, extracted from the summary. Our metric aligns claims from the summary with text from the input at three different levels of granularity: sentence, paragraph, and document. 
This novel alignment strategy facilitates interpretability: when presented with a claim, the metric user can either verify which specific part of the input document entails it, or, in the presence of hallucinations, they can observe that no alignment is provided by the metric.

Not only does our approach enhance interpretability but it also yields a positive impact on overall performance: \metric~achieves the highest average result in the AggreFact benchmark \citep{tang-etal-2023-understanding}, the de-facto benchmark for summarization factuality evaluation. 
Furthermore, we evaluate the performance of our metric in a more challenging context, specifically targeting long-form summarization. To accomplish this, we conduct a human annotation process aimed at assessing long-form summarization factuality. When confronting the metrics outputs with the human labels, {\metric} exhibits higher accuracy compared to its competitors, paving the way for future approaches for long document summarization evaluation.  

In summary, our contributions are the following:
\begin{itemize}
    \item We introduce \metric, a state-of-the-art factuality metric for summarization evaluation;
    \item The alignments provided by our metric facilitate interpretability, highlighting the input document sections that are useful for summary claims verification;
    \item {\metric} is designed to be computationally efficient, utilizing approximately 700 million parameters, thereby ensuring its applicability in real-world applications. 
    \item We show the effectiveness of {\metric} in the evaluation of long-form summarization factuality through a human annotation process.
    \item In the hope of fostering research in summarization factuality evaluation, we release the code of our metric and our human annotations at \url{https://github.com/Babelscape/FENICE}.
\end{itemize}

\section{Related Work}

\paragraph{NLI-based metrics.}

Drawing upon recent developments in the NLI field, several studies have introduced novel factuality metrics that exhibit stronger correlations with human judgments. \citet{chen2023menli} proposed a series of NLI metrics tailored for both MT and summarization evaluation.
They conducted experiments in both reference-based and reference-less settings. In the latter setting, which serves as the standard for consistency evaluation, they utilized the input document as the premise for the NLI system and the entire summary as the hypothesis for verification. While this approach correlates well with human consistency judgments, it lacks interpretability: on the one hand, when a summary is inconsistent, the end user is not provided with the parts of it that are deemed non-factual; on the other hand, when a summary is factually accurate, the metric provides no indication of the specific sections in the document that entail it.

In the same context, AlignScore \citep{zha-etal-2023-alignscore} provides a novel alignment system, which is trained on several verification tasks, including NLI. While it achieves high performance in factuality evaluation, AlignScore also falls short in terms of interpretability, providing an overall score for the entire summary, which prevents users from identifying specific factual or non-factual spans within it. Moreover, it aligns the summary with large blocks of text from the input document, making it hard for users to gain insights from the output.

The Dependency Arc Entailment~\cite[DAE]{goyal-durrett-2020-evaluating} metrics leverage the entailment between the dependency parse trees of document sentences and the summary. However, the highly abstractive nature of summaries, often encapsulating multiple facts in a single sentence, introduces a drawback, making their sentence-level comparison suboptimal.

Despite the significant progress achieved so far, we posit that there is still considerable room for enhancing NLI-based metrics, encompassing improvements in both performance and interpretability.

\paragraph{LLM-based metrics.}
Several studies \citep{shen2023large, chen2023evaluating, wang2023chatgpt} have recently analyzed the efficacy of LLMs as summarization evaluators in multiple evaluation settings. Their experiments demonstrate that, while LLMs outperform existing automatic metrics, they are not yet ready to be considered reliable alternatives for human evaluation.
Starting from these findings, TrueTeacher~\cite{gekhman2023trueteacher} analyzed the LLMs' capability of generating large-scale factuality datasets and then trained smaller student models on such data, outperforming existing NLI metrics.  

Another trend of approaches revolves around prompt engineering strategies~\cite{luo2023chatgpt,fu2023gptscore,liu2023geval, gao2023humanlike}, where LLMs are typically instructed with the evaluation task description, the input document, and the system-generated summary. Specifically, \citet{fu2023gptscore}~demonstrated the capabilities of such models in achieving multi-aspect, customized, and training-free evaluation. Relatedly, the recently-introduced G-Eval~\citep{liu2023geval} experimented with chain-of-thoughts (CoT) and form-filling instruction paradigms. They showed that G-Eval, configured with GPT-4, outperforms all previous methods in terms of correlation with the human consistency annotations in SummEval~\citep{fabbri-etal-2021-summeval}. 

While achieving high correlations with human judgments, LLM-based metrics raise concerns about their real-world applicability. Firstly, many of these studies rely on closed-source LLMs, and the lack of information about the composition of their training dataset raises uncertainty: there is no guarantee that the test examples used in these studies are actually unseen at training time for these models. Additionally, their exceptionally-high computational cost poses a significant constraint on their practical deployment and hampers their reproducibility for research purposes.

Our approach differs from previous ones in that it utilizes an LLM solely for the task of claim extraction, rather than also utilizing it for generating the factuality annotations, which it may have already seen during training. Furthermore, we provide a distilled version of this LLM—a more compact generative Transformer—trained to emulate the output of the LLM, which ensures the practicality and reproducibility of our approach.

\paragraph{Claim extraction-based
evaluation.} 
A growing number of works explores the employment of claim extraction for the purpose of evaluating generated text. Among these, ~\citet{min2023factscore} proposed a method for evaluating long-form text generation, through the validation of their extracted claims given a reliable knowledge source, e.g., Wikipedia.

In the context of summarization evaluation, ~\citet{pyramid} introduced Pyramid, the first manual annotation protocol in this context. However, the scope of Pyramid and its automatic approximations~\cite{gao-etal-2019-automated, harnly-pyramid} is to evaluate summarization relevance, which differs significantly from the factuality-oriented focus of our study.

To the best of our knowledge, we are the first to consider claims as the foundation for a factuality-oriented summarization
metric.

\section{\metric}
We now provide an overview of the {\metric} algorithm. Initially, we extract a list of atomic facts, referred to as claims, from a given summary (Section \ref{sec:claim_extraction}). Subsequently, we detail the process of obtaining a factuality score for each extracted claim. This involves leveraging NLI, with the claim serving as the hypothesis, and the document sentences (Section \ref{sec:nli_score}) or longer blocks of text at different levels of granularity (Section \ref{sec:multi_grained}) serving as premises. In Section \ref{sec:coreference}, we elaborate on a refinement of the obtained scores through coreference resolution.

\subsection{Claim extraction}
\label{sec:claim_extraction}
Following the Atomic Content Unit (ACU) definition outlined in \citet{liu2023revisiting}, we use the term \textit{claim} to denote an \textit{atomic fact}, i.e., an elementary information unit found in a summary, that does not require further subdivision. Figure~\ref{fig:overview}(a) depicts an example list of claims extracted from a summary.

We frame the claim extraction problem as an end-to-end autoregressive generation task. Given a prompt consisting of the task description and the summary to be processed, the model generates a sequence of claims expressed within the provided summary. To guide the model in meeting the desired output format and enhancing its task comprehension, we include an example of an input summary and the expected claims in the prompt (see Appendix \ref{app:prompt}).

We opt to center our evaluation strategy around claims rather than sentences, primarily because of their enhanced interpretability and their atomic nature, which facilitates alignment with information from the input.
\subsection{NLI-based claim scoring}
\label{sec:nli_score}
The NLI formulation lies at the core of our novel claim scoring algorithm. NLI is a task that involves determining the logical relationship between two text segments: a \textit{premise} and a \textit{hypothesis}. Formally, given a premise $pre$ and a hypothesis $hyp$:
\begin{equation}
\label{eq:nli_formulation}
\text{NLI}(pre, hyp) = (p_{\text{ent}}, p_{\text{neu}}, p_{\text{con}})
\end{equation}
where $p_{ent}$, $p_{neu}$, and $p_{con}$ denote the respective probabilities of $pre$ entailing, being neutral about, or contradicting $hyp$.  

Starting from Equation \ref{eq:nli_formulation}, we define $\text{NLIScore}$ as the difference between the entailment and contradiction probabilities, obtained using the document sentence as premise and the claim extracted from the summary as hypothesis:
\begin{align*}
\text{NLIScore}(s, c) = p_{\text{ent}} - p_{\text{con}}
\end{align*}
where the premise $s$ is a sentence of the input document, and the hypothesis $c$ is a claim extracted from the summary.
NLIScore measures the degree to which a specific premise can be utilized to verify the claim $c$. The inclusion of the contradiction probability serves the purpose of penalizing claims that are in conflict with any content in the source document.

Finally, we introduce $\text{\metric}_{sent}$ as the function that assigns a score to a given claim $c$ by considering the sentences in the  source document $D$ as premises: 
%
\begin{equation}
\label{eqn:fenice_sent}
\text{\metric}_{sent}(D, c) = \max\limits_{s \in D} \text{NLIScore}(s, c)\\
\end{equation}

We also define $\text{align}_{sent}$ as the corresponding alignment function: 
\begin{equation}
\label{eqn:align_sent}
\text{align}_{sent}(D, c) = \argmax\limits_{s \in D} \text{NLIScore}(s, c)\\
\end{equation}
For each claim $c$ from the summary, this establishes a one-to-one alignment between $c$ and the highest-scoring sentence from the input document.

\subsection{Coreference Resolution}
\label{sec:coreference}

Importantly, given a claim $c$ and its aligned sentence $s$ = \text{align}$_{sent}(D, c)$, NLIScore$(s, c)$ may not reflect the true degree of entailment for the $(s, c)$ pair due to an entity or concept being mentioned in the two texts with distinct surface forms. 
%
%
For instance, consider the claim in Figure \ref{fig:overview}(a): "\textit{England number eight Billy Vunipola is set to return from injury}" and its aligned source document sentence: "\textit{There were initial fears the 24-years-old player would miss} [...]". In this case, the NLI model assigns only a moderate score of 0.6 to this pair due to the limited context, making it challenging to establish a clear correspondence between "Billy Vunipola" and "the 24-years-old player".

To tackle this challenge, we apply a Coreference Resolution model on the source document, which identifies clusters of mentions that refer to the same entity or concept. We then introduce a coreference-resolution-aware version of Equation~\ref{eqn:fenice_sent} as follows:
%
\begin{equation}
\begin{aligned}
s^* & = \text{align}_{sent}(D, c)\\
\text{\text{\metric}}_{\text{coref}}(D, c) & = \max_{s' \in \text{Co}(s^*)} \text{NLIScore}(s', c)
\end{aligned}
\end{equation}

\noindent where $Co(s^*)$ is the set of sentences obtained by replacing each entity occurring in sentence $s^*$ with all its coreferential mentions.\footnote{We limit this augmentation to the sentence $s^*$, which is the result of Equation \ref{eqn:align_sent}.}

For instance, given "the 24-years-old player" we identify its coreferential spans in the document, such as "Billy Vunipola". Subsequently, we generate an alternative version of the original sentence by replacing "the 24-years-old player" with "Billy Vunipola", resulting in the sentence: \textit{"There were initial fears Billy Vunipola would miss} [...]". As depicted in Figure~\ref{fig:overview}(b), this approach facilitates the alignment process of the claim \textit{"England number eight Billy Vunipola is set to return from injury"} to the revisited version of the sentence, which contains the same span for the given named entity.

\subsection{Aligning claims across multiple input text granularities}
\label{sec:multi_grained}
Given the abstract nature of summaries, the verification of a claim may require a context that extends beyond a single sentence from the input. For example, to verify the claim: \textit{"Regular exercise contributes to overall well-being"}, the following input passage would be necessary: \textit{"Studies have consistently shown that individuals who engage in regular physical activity experience improved mental health. Furthermore, regular exercise is linked to a reduced risk of chronic diseases and enhanced cardiovascular health."}. 

Therefore, we define the following variant of Equation \ref{eqn:fenice_sent} for longer passages of $k$ sentences:
\begin{equation}
\small
\label{eqn:fenice_passage}
\text{\metric}_{p}(D, c, k) = \max\limits_{i=1, \dots, |D|-k+1} \text{NLIScore}(s_{i}^{i+k-1}, c)\\
\end{equation}

\noindent where $s_{i}^{i+k-1}$ is the concatenation of $s_i, \dots, s_{i+k-1}$, i.e., the $k$ consecutive sentences starting from sentence $i$ in document $D$.

Our approach involves choosing $k$ such that $k < |D|$, alongside considering the entire document as premise, i.e., $k=|D|$.
Thus, our function for scoring claims given passages of multiple lengths from the source document is given by the following:
\begin{equation}
\small
\label{eqn:fenice_multi}
\begin{split}
\text{\metric}_{mul}(D, c, j) = \max\{\text{\metric}_{p}(D, c, j), \\\text{\metric}_{p}(D, c, |D|)\}\\
\end{split}
\end{equation}
where $j$ is the desired length for the contiguous blocks of the source sentences.
Then, we define $\text{\metric}_{cl}$ as the factuality score assigned to a claim $c$:
\begin{equation}
\small
\begin{split}
\text{\metric}_{cl}(D, c, j, T) =\left\{
\begin{aligned}
    &\text{\metric}_{mul}(D, c, j)\\
    &\quad \text{if } \text{\metric}_{coref}(D, c) < T\text{;}\\
    &\text{\metric}_{coref}(D, c) \quad \text{otherwise.}
\end{aligned}
\right.
\end{split}
\end{equation}
where $T$ is a threshold. We introduce $T$ to limit the application of $\text{\metric}_{mul}$ to only a subset of claims that requires further verification, i.e., the low-scoring ones. We set $j=5$ and $T=0.8$ in all our experiments (Section~\ref{sec:factuality_evaluation}) as a result of tuning them on the validation splits.

Finally, given the set of extracted claims $C$ from a summary $S$, and a source document $D$, we introduce $\text{\metric}$ as:
\begin{equation}
\small
\label{eqn:fenice}
\text{\metric}(D, C, j, T) = \frac{\sum_{c\in C}\text{\metric}_{cl}(D, c, j, T)}{|C|}
\end{equation}

\noindent which is the factual consistency score attributed by our metric to summary $S$.

\section{Experiments and Results}
\subsection{Claim extraction}
\paragraph{Experimental setup.}
As detailed in Section \ref{sec:claim_extraction}, we formulate the claim extraction problem as a sequence generation task. To accomplish this, we designed a prompt that includes the task description, an input-output example, and the summary to be processed. The prompt utilized is available in Appendix \ref{app:prompt}.
We experiment with \textit{GPT-3.5-turbo} as claim extractor.\footnote{We use the \textit{gpt-3.5-turbo} API, version dated 2nd December 2023.}
We additionally provide a distilled version of the claim extractor, which consists of a T5$_{base}$~\citep{T5} model trained on the LLM outputs. 
To create the training and evaluation datasets for the distilled model, we apply the LLM-based claim extractor on the reference summaries from two widely-used summarization datasets, namely XSum~\citep{narayan-etal-2018-dont} and CNN/DailyMail~\citep{cnn-daily-mail}. We split the dataset obtained into training, validation, and test sets following a 80\%-10\%-10\% proportion, comprising 430,946 summary-claims pairs for training, and 53,868 for both the validation and test partitions. We train our system on a single GPU \textit{NVIDIA GeForce RTX 3090} for a total of 1M steps, using AdaFactor~\citep{adafactor} as optimizer with learning rate of $1\cdot10^-5$.
\paragraph{Datasets.}
We evaluate our distilled version of the claim extractor on the test split of the LLM-generated claim extraction dataset.
We additionally assess the performance of both the LLM-based claim extractor and our distilled version on ROSE~\citep{liu2023revisiting}, a dataset that includes summaries and human-written claims among its annotations. While this dataset was designed to serve as the basis for a human annotation protocol for summarization evaluation, we utilize the released manually-extracted claims for assessing our systems' performance.

\paragraph{Metrics.}
Given the extractive nature of this task, we consider ROUGE~\citep{lin-2004-rouge} as the basis for a suitable evaluation metric for claim extraction. Following \cite{zhang-bansal-2021-finding} we define $\text{easiness}_P$ as the score obtained by computing the ROUGE-1 metric for each pair of system-generated and gold standard claims, subsequently calculating the maximum, and averaging over the total number of generated claims:
\begin{equation}
\label{eq:easiness_p}
    \text{easiness}_P(S, H) = \frac{\sum_{c\in S}\max_{c^* \in H}\text{R1}(c, c^*)}{|S|}
\end{equation}
where $R1$ is the ROUGE-1-F1 metric, and $S$, $H$ are the sets of system- and human-written claims, respectively.
We then introduce $\text{easiness}_R$ as the recall-oriented version of Equation~\ref{eq:easiness_p}, obtained by averaging over the gold standard claims:
\begin{equation}
\label{eq:easiness_r}
    \text{easiness}_R(S, H) = \frac{\sum_{c^* \in H}\max_{c\in S}{\text{R1}(c, c^*)}}{|H|}    
\end{equation}
\paragraph{Results and discussion.}
In Table~\ref{tab:claim_extraction_rose} we report the results of the $\text{easiness}_{P}$, and $\text{easiness}_{R}$ metrics, along with the F1 version of Equations \ref{eq:easiness_p} and \ref{eq:easiness_r}, i.e., $\text{easiness}_{F1}$, obtained on ROSE. In Appendix~\ref{app:transformer_claims} we additionally report the results obtained when using BERTScore~\cite{zhang2020bertscore} as backbone metric for our evaluation instead of ROUGE.
\begin{table}[t]
    \centering
    \begin{tabular}{lccc}
        \toprule
        \textbf{Model} & \textbf{$\text{easiness}_P$} & \textbf{$\text{easiness}_R$} & \textbf{$\text{easiness}_{F1}$}\\
        \midrule
        GPT-3.5 & 80.1 & 70.9 & 74.9 \\
        T5$_\textit{dist\_GPT}$ & 79.2 & 68.8 & 73.4 \\ 
        \bottomrule
    \end{tabular}
    \caption{Easiness Precision ($\text{easiness}_P$), Recall ($\text{easiness}_R$), and F1 score ($\text{easiness}_{F1}$) results for the LLM-based claim extractor, namely \textit{GPT-3.5}, and its distilled version \textit{T5$_\textit{dist\_GPT}$}, assessed on ROSE~\citep{liu2023revisiting}.}
    \label{tab:claim_extraction_rose}
\end{table}

\begin{table}[t]
    \centering
    \begin{tabular}{lccc}
        \toprule
        \textbf{Model} & \textbf{$\text{easiness}_P$} & \textbf{$\text{easiness}_R$} & \textbf{$\text{easiness}_{F1}$}\\
        \midrule
        T5$_\textit{dist\_GPT}$ & 89.7 & 89.9 & 89.5 \\
        \bottomrule
    \end{tabular}
    \caption{$\text{Easiness}$ metric results of the distilled version of the LLM-based claim extractor, i.e., \textit{T5$_\textit{dist\_GPT}$ } evaluated on the test set of the dataset composed of claims generated by \textit{GPT-3.5}.}
    \label{tab:claim_extraction_distill}
\end{table}

As we can see from Table~\ref{tab:claim_extraction_rose}, our distilled version performs comparably with the original one based on \textit{GPT-3.5}. These results indicate the effectiveness of the distillation process, ensuring that our model can be used reliably for extracting claims. To further evaluate the quality of the claims generated by our distilled model, we compare them to those generated by the original GPT-3.5 model, using the latter as references.
The results of this evaluation are presented in Table~\ref{tab:claim_extraction_distill}, demonstrating the strong ability of our T5 model to replicate the claims generated by GPT-3.5.

\subsection{Factuality Evaluation}
\label{sec:factuality_evaluation}
\paragraph{Experimental setting.}
We leverage an off-the-shelf model trained on NLI to align the summary-extracted claims with text from the input document. Specifically, we employ a version of DeBERTa-v3-large~\citep{deberta} fine-tuned on multiple NLI datasets,\footnote{We loaded the pre-trained model weights from the Hugging Face Hub, available at: \url{https://huggingface.co/MoritzLaurer/DeBERTa-v3-large-mnli-fever-anli-ling-wanli}} comprising MultiNLI~\cite{mnli}, LingNLI~\citep{parrish-etal-2021-putting-linguist}, WANLI~\citep{liu-etal-2022-wanli}, and Fever-NLI, as well as Adversarial-NLI, both introduced in ~\citet{fever-nli}. We adopt spaCy's \textit{en-core-web-sm} to segment the input documents into sentences, and F-Coref~\citep{otmazgin-etal-2022-f} for the extraction of the coreferential spans (Section~\ref{sec:coreference}).

\paragraph{Evaluation datasets.}
We assess the effectiveness of \metric~on \textsc{AggreFact}~\citep{tang-etal-2023-understanding}, the de-facto benchmark for evaluating summarization factuality metrics. \textsc{AggreFact} comprises multiple datasets of news articles and manually annotated summaries with factuality labels. The benchmark is categorized into three splits based on the summarization systems employed: i) \textsc{FtSota}, which includes the outputs of state-of-the-art pretrained summarizers, 
ii) \textsc{ExFormer}, consisting of summaries of early Transformer-based summarization models
, and iii) \textsc{Old}, which comprises the outputs of older approaches. 


\paragraph{Baselines.}
We evaluate the performance of \metric~against several baselines that align with recent trends in the field, specifically focusing on NLI, QA, and LLM-based metrics.

\subparagraph{NLI-based metrics.}
We evaluate several NLI-based metrics, such as AlignScore~\cite{zha-etal-2023-alignscore}, MENLI~\cite{chen2023menli}, SummaC-ZS and SummaC-Conv, both introduced in \cite{laban-etal-2022-summac}, and DAE \citep{goyal-durrett-2020-evaluating}.
\begin{table}[t]
\resizebox{\linewidth}{!}
{
\centering
\setlength\tabcolsep{3pt}  
\begin{tabular}{lccc|ccc|c}
\toprule
 & \multicolumn{3}{c|}{\textbf{Agg-CNN}} & \multicolumn{3}{c|}{\textbf{Agg-XSum}} & \multicolumn{1}{c}{} \\
 & \textbf{\textsc{FtS}} & \textbf{\textsc{ExF}} & \textbf{\textsc{Old}} & \textbf{\textsc{FtS}} & \textbf{\textsc{ExF}} & \textbf{\textsc{Old}} & \textbf{AVG} \\
\midrule
Random Baseline & 50.0 & 50.0 & 50.0 & 50.0 & 50.0 & 50.0 & 50.0 \\
\midrule
DAE* & 59.4 & 67.9 & 69.7 & 73.1 & - & - & 67.5 \\
QuestEval & 63.7 & 64.3 & 65.2 & 61.6 & 60.1 & 59.7 & 63.5 \\
SummaC-ZS & 63.3 & \textbf{76.5} & 76.3 & 56.1 & 51.4 & 53.3 & 65.3 \\
SummaC-Cv & \textbf{70.3} & \underline{69.8} & 78.9 & 67.0 & 64.6 & 67.5 & 70.5 \\
TrueTeacher-11B & 65.7 & 57.7 & \underline{81.9} & \underline{75.2} & 68.4 & 52.8 & 67.0 \\
QAFactEval & 61.6 & 69.1 & 80.3 & 65.9 & 59.6 & 60.5 & 67.0 \\
MENLI & 51.7 & 52.8 & 68.4 & 58.3 & 59.7 & \textbf{73.9} & 60.8 \\
AlignScore & 53.5 & 73.9 & 78.0 & \textbf{80.2} & \textbf{79.9} & 63.7 & \underline{71.5} \\
\midrule
ChatGPT-ZS & 66.2 & 64.5 & 74.3 & 62.6 & 69.2 & 60.1 & 65.7 \\
ChatGPT-CoT & 49.7 & 60.4 & 66.7 & 56.0 & 60.9 & 50.1 & 57.9 \\
ChatGPT-DA & 48.0 & 63.6 & 71.0 & 53.6 & 65.6 & 61.5 & 60.1 \\
ChatGPT-Star & 55.8 & 65.8 & 71.2 & 57.7 & 70.6 & 53.8  & 63.8 \\
\midrule
\text{\metric}$_{GPT\_claims}$ & \underline{68.2} & 68.8 & \textbf{82.1} & 73.9 & \underline{73.5} & \underline{69.9} & \textbf{72.7} \\
\text{\metric}$_{T5\_claims}$ & 61.6 & 67.8 & 80.6 & 71.4 & 70.7 & 67.7 & 70.0 \\
\bottomrule
\end{tabular}
}
\caption{Balanced accuracy results on the test sets of the \textsc{AggreFact} datasets. Following ~\citet{tang-etal-2023-understanding} we exclude the performance of DAE on the \textsc{ExFormer} (\textsc{ExF} in table)
and \textsc{Old} datasets in the \textsc{AggreFact-XSum} partition, since it was trained on XSumFaith~\cite{goyal-durrett-2021-annotating} which is part of those splits. Results in \textbf{bold} indicate the best performance, while \underline{underlined} values indicate the second best.}
\label{tab:aggrefact}
\end{table}

\subparagraph{QA-based metrics.}
Question Answering-based factuality metrics typically depend on a Question Generation (QG) model designed to formulate questions based on the input document as context, and check whether information in the summary can be used to answer such questions, and vice versa. For our study, we choose the top-performing QA metric on \textsc{AggreFact} as baseline, namely QAFactEval~\citep{fabbri-etal-2022-qafacteval}.

\subparagraph{LLM-based metrics.}
We benchmark recent ChatGPT-based evaluation metrics: ChatGPT-ZS and ChatGPT-CoT \cite{luo2023chatgpt} which leverage GPT-3.5 to generate a binary factuality label, in zero-shot and Chain-of-Thought paradigms, respectively. We also include ChatGPT-DA and ChatGPT-Star \cite{wang2023chatgpt}, which prompt LLMs to generate factuality scores on a scale of 0-100 and 1-5, respectively. We additionally include the results of the 11B-parameter T5 version of the TrueTeacher metric.
\begin{table}[t!]
    \centering
    \resizebox{\linewidth}{!}{%
        \setlength\tabcolsep{3pt}  
        \begin{tabular}{lcc|c}
            \toprule
            \textbf{Metric} & \textsc{\textbf{CNN(FtS)}} & \textsc{\textbf{XSum(FtS)}} & \textsc{\textbf{AVG}} \\
            \midrule
            DAE & 65.4$\pm$4.4 & 70.2$\pm$2.3 & 67.8\\
            QuestEval & \underline{70.2}$\pm$3.2 & 59.5$\pm$2.7 & 64.9\\
            SummaC-ZS & 64.0$\pm$3.8 & 56.4$\pm$1.2 & 60.2\\
            SummaC-Conv & 61.0$\pm$3.9 & 65.0$\pm$2.2 & 63.0\\
            QAFactEval & 67.8 $\pm$4.1 & 63.9$\pm$2.4 & 65.9\\
            TrueTeacher-11B & 62.0$\pm$1.3 & \textbf{74.9}$\pm$1.2 & 68.4\\
            MENLI & 63.4$\pm$2.8 & 59.0$\pm$1.8 & 61.2\\
            AlignScore & 62.7$\pm$3.1 & 69.4$\pm$1.9 & 66.1\\ 
            \midrule
            ChatGPT-ZS & 56.3$\pm$2.9 & 62.7$\pm$1.7 & 59.5 \\
            ChatGPT-COT & 52.5$\pm$3.3 & 55.9$\pm$2.1 & 54.2\\
            ChatGPT-DA & 53.7$\pm$3.5 & 54.9$\pm$1.9 & 54.3\\
            ChatGPT-Star & 56.3$\pm$3.1 & 57.8$\pm$0.2 & 57.1\\
            \midrule 
            \text{\metric$_{GPT\_claims}$} & \textbf{70.5}$\pm$1.8 & \underline{72.8}$\pm$0.3 & \textbf{71.6}\\
            \text{\metric$_{T5\_claims}$} & 67.7$\pm$3.0 & 70.1$\pm$1.8 & \underline{68.9} \\
            \bottomrule
        \end{tabular}%
    }
    \caption{Balanced accuracy results on the \textsc{AggreFact-FtSota} split obtained with the single-threshold setting. Results in \textbf{bold} indicate the best performance, while \underline{underlined} values represent the second best.}
\label{tab:aggrefact_singlet}

\end{table}

\paragraph{Results and discussion.}
We assess the performance of two distinct variants of our metric: \metric$_{GPT\_claims}$, which employs LLM-based claims, and \metric$_{T5\_claims}$, featuring the knowledge-distilled claim extractor.
To convert \metric~scores into binary labels, we tune a different threshold on each validation split of the \textsc{AggreFact} datasets (i.e., \textsc{FtSota}, \textsc{ExFormer}, and \textsc{Old}) and then compute the balanced accuracy on the corresponding test set, as indicated in \citet{tang-etal-2023-understanding}.

In Table~\ref{tab:aggrefact}, we present the balanced accuracy~\cite{bal_accuracy} scores on the \textsc{AggreFact} test sets. \text{\metric$_{GPT\_claims}$}~sets a new state-of-the-art performance on \textsc{AggreFact}, resulting in the highest average balanced accuracy score.

Moreover, we specifically analyze the metric's ability to evaluate the outputs of state-of-the-art models. Following the recommendation of \citet{tang-etal-2023-understanding}, we report the results of our metrics on \textsc{AggreFact-FtSota} split in the single-threshold setting. As outlined by \citet{tang-etal-2023-understanding}, we fine-tune a threshold on the validation sets of the \textsc{CNN-FtSota} and \textsc{XSum-FtSota} splits and subsequently apply it to convert our metric scores into binary labels on the test set.
From the outcomes presented in Table~\ref{tab:aggrefact_singlet}, we note that \metric$_{GPT\_claims}$ obtains the best average results in this split, also due to its high performance on \textsc{XSum-FtSota}. Notably, we consider this latter setting to be the most challenging for a metric, given that it comprises summaries of state-of-the-art systems trained on XSum references, which exhibit the highest level of abstraction.
In this context, \metric$_{T5\_claims}$ slightly outperforms TrueTeacher, despite utilizing significantly fewer parameters (700M vs 11B).
\subsection{Ablation study}
\label{app:ablation}
We conduct an analysis of the impact of our primary design choices, namely, claim extraction, coreference resolution, the alignment with multiple text granularities, and our backbone NLI model. 

First, we introduce NLI$_{sent}$, a simple baseline leveraging NLI-based sentence-level alignment between the summary and the input document. Specifically, the factuality score for a summary sentence $s'$ is obtained through the following variant of Equation~\ref{eqn:fenice_sent}, considering summary sentences as hypotheses instead of claims:
\begin{equation}
\label{eqn:fenice_sent_sent}
\text{NLI}_{sent}(D, s') = \max\limits_{s \in D} \text{NLIScore}(s, s')\\
\end{equation}

\noindent As outlined in Section~\ref{sec:multi_grained}, we obtain the factuality score for a summary by averaging the result of Equation \ref{eqn:fenice_sent_sent} applied to its sentences. 

We then assess the impact of focusing the factuality evaluation on claims extracted from the summary rather than its sentences.
To achieve this, we introduce NLI$_{claim}$, a modification of the previous baseline that assigns factuality scores to the claims using Equation~\ref{eqn:fenice_sent}.
Subsequently, we evaluate the performance of NLI$_{coref}$, derived by integrating the coreference resolution strategy (Section~\ref{sec:coreference}) into the previous baseline. Additionally, we assess the impact of including the alignment strategy with input text at multiple granularities (Section~\ref{sec:multi_grained}) by presenting the performance of \metric. Table~\ref{tab:ablation} displays the balanced accuracy scores of \metric~on the \textsc{AggreFact- FtSota} split, comparing them to the aforementioned baselines.
\begin{table}[t!]
    \centering
    \small
    \setlength\tabcolsep{3pt}  

    \begin{tabular}{lcccc}
        \toprule
        \textbf{Metric} & \textbf{\textsc{CNN}} & \textbf{\textsc{XSum}} & \textbf{AVG} & \textbf{$\Delta$ AVG} \\
        \midrule
        
        NLI$_{sent}$ & 52.2$\pm$1.4	& 59.6$\pm$0.6	& 55.9 & ~~\_ \\
        NLI$_{claim}$ & 54.3$\pm$2.1 & 63.2$\pm$0.8	& 58.8 & ~~+2.9 \\
        NLI$_{coref}$ & 59.7$\pm$2.3& 63.7$\pm$0.9	& 61.5	& ~~+2.7 \\
        \metric & \textbf{70.5}$\pm$3.0 & \textbf{72.8}$\pm$0.3 & \textbf{71.6} & \textbf{+10.1} \\

        \bottomrule
    \end{tabular}
   \caption{Ablation study evaluating the impact of component additions on \metric~on the \textsc{AggreFact-FtSota} split. We report the results of the following baselines: i) NLI$_{sent}$, utilizing summary sentences instead of claims, ii) NLI$_{claim}$, which leverages claims, and iii) NLI$_{coref}$, obtained by integrating the coreference-resolution-based alignments.}

    \label{tab:ablation}
\end{table}
The values in the $\Delta AVG$ column represent the differences in terms of average balanced accuracy resulting from the incremental inclusion of each component.
The results indicate that each component contributes significantly to the final performance. The alignment strategy with input text at multiple granularity has the most substantial impact, resulting in a significant gain of 10.1 points when included. Additionally, we evaluate the performance of \metric~utilizing RoBERTa as the NLI model for our alignments (\metric$_{RoBERTa}$), instead of DeBERTa (Appendix~\ref{app:nli_ablation}).



\subsection{Long-form summarization evaluation}
In recent years, there has been a noticeable shift in research focus away from summarizing short texts, predominantly news articles, to handling longer documents like scientific papers~\citep{gupta-etal-2021-sumpubmed} and books~\cite{kryściński2022booksum,scire-etal-2023-echoes}. Despite this rapid evolution in the development of approaches and resources, the evaluation of summarization has been slower to adapt, with only a limited number of works, which are primarily aimed at defining the basic guidelines for this task~\citep{krishna-etal-2023-longeval}.

\paragraph{Human evaluation.}
As an initial step towards addressing this gap, we conducted an evaluation of our metric in the context of long-form text summarization. To do so, we curated a set of annotations for assessing summaries of lengthy texts.

Drawing upon Echoes from Alexandria~\cite{scire-etal-2023-echoes}, a multilingual resource for book summarization, we selected 26 English narratives, each paired with a reference abstractive summary.
The reference summaries were refined by removing irrelevant details such as the story title, the genre, and the author's name. As an example, the initial summary \textit{"Isaac Asimov's 'Let's Get Together' is a science fiction novel which depicts the continuation of Cold War hostilities..."} underwent manual editing and was refined to \textit{"The story depicts the continuation of Cold War hostilities..."}.
To augment our selection, we incorporated summaries generated by 12 book summarization systems, as presented by \citet{scire-etal-2023-echoes}, applied to our selected stories. The systems chosen were versions of BART~\citep{lewis-etal-2020-bart}, LongT5~\citep{guo-etal-2022-longt5}, and LED~\citep{led}, fine-tuned on the \textit{Echo-XSum} split of the \textit{Echoes from Alexandria} dataset. This process resulted in a comprehensive set of 338 text-summary pairs.

Next, we tasked an expert annotator with reading all the stories and assigning a binary factuality label to each of the associated summaries.\footnote{We supplied the annotator with the PDF files containing the text of the stories. We paid the annotator according to the standard salary for their geographical location. The annotator took an average of 2 hours to read a story, and an average of 5 minutes to annotate the corresponding summaries.}
As the foundation for the annotation process, we considered a summary  factual if its content was explicitly stated or implied by the source document; otherwise, it was labeled as non factual.
Through this process, we acquired factuality labels for all the 338 text-summary pairs.

Moreover, in order to measure the inter-annotator agreement we asked a second expert to manually annotate a subset of 52 summaries. We then computed the Cohen's kappa coefficient, obtaining a moderate agreement of 0.74. We deemed this outcome reasonable, particularly when considering the high complexity of assessing long-form summarization compared to the standard setting.
\begin{table}[t]
    \centering
    \begin{tabular}{ccclll}
        \toprule
        \textbf{Dataset} & \textbf{Avg. Sentences} & \textbf{Avg. Tokens}\\
        \midrule
        Ours & 664.2 & 11,817.9  \\
        SummEval & ~~18.0 & ~~~~~363.1 \\
        SummaC & ~~23.0 & ~~~~~485.5 \\
        \bottomrule
    \end{tabular}
    \caption{Our human annotations for long-form text summarization statistics compared to standard-form (news) factuality benchmarks, i.e., SummaC and SummEval.}
\label{tab:fairy_summeval_stats}
\end{table}
Table~\ref{tab:fairy_summeval_stats} compares the length of source texts in our proposed long-form summarization evaluation dataset with those in existing datasets, specifically SummaC~\cite{laban-etal-2022-summac} and SummEval~\cite{fabbri-etal-2021-summeval}. In our dataset, annotators were tasked with reading very long source texts, averaging 664 sentences, in order to assess the factuality of the provided summaries. The distribution of factual and unfactual labels in our dataset is detailed in Appendix~\ref{app:fairy_summeval_dist}.
To align with the evaluation setup of \textsc{AggreFact}, we partitioned our annotations into test and validation sets using an 80/20 split.   
 
\begin{table}[t!]
    \centering
    \small

    \begin{tabular}{lc}
        \toprule
        \textbf{Metric} & \textbf{Balanced Accuracy} \\
        \midrule
        \text{AlignScore} & 61.3 \\
        \text{MENLI} & 61.7 \\
        \text{DAE} & 51.4 \\
        \text{\metric$_{GPT\_claims}$} & \textbf{66.2} \\
        \text{\metric$_{T5\_claims}$} & \underline{65.7} \\

        \bottomrule
    \end{tabular}
    \caption{Balanced accuracy results of \metric~ metrics and the selected baselines on our set of human annotations for long form summarization factuality. Results in \textbf{bold} indicate the best performance, while \underline{underlined} values represent the second best.}
    \label{tab:long_eval}
\end{table}
\paragraph{Results and discussion.}
We follow the experimental setup outlined in Section~\ref{sec:factuality_evaluation}. We employ AlignScore, MENLI, and DAE as baselines, considering their relatedness to our work. In Table~\ref{tab:long_eval}, we present the balanced accuracy results on the test split of our dataset.
The results obtained highlight that \metric~metrics outperform the baselines, with \text{\metric$_{GPT\_claims}$} obtaining the highest performance and \text{\metric$_{T5\_claims}$} coming second. Upon manual inspection of our alignments, we observe that a significant proportion of claims are aligned with source text having the length of a paragraph. We posit that this behaviour is due to the high abstractiveness of the stories' summaries, which typically require contexts that are longer than a sentence in order to be verified. We hypothesize that our metric is well-suited for both standard and long-form summarization thanks to its adaptive capacity to choose the appropriate input text granularity for claim verification.

\section{Conclusion and Future Work}
We introduced \metric, a novel metric designed for the factuality evaluation of summarization. By leveraging NLI-based alignments between claims extracted from the summary and the source text at multiple levels of granularities, {\metric} facilitates interpretability. Our experiments demonstrate that our metric attains state-of-the-art performance in standard summarization evaluation datasets. Additionally, we showcase the effectiveness of our metric in a more-challenging setting, i.e., the evaluation of long-form summarization. To do so, we conducted a human annotation process for this task. Our work paves the way for future research focused on assessing long-form summarization factuality. As future work, we plan to improve our zero-shot alignment system's performance with a tailored fine-tuning on summarization datasets.

\clearpage
\label{app:conclusion}
\section*{Limitations}
While our approach incorporates multiple components, including NLI, coreference resolution and claim extraction, in order to enhance the robustness and comprehensiveness of the factuality evaluation metric, it is essential to acknowledge that the involvement of these diverse systems introduces a layer of complexity and potential sources of error. The inherent variability and imperfections in each of these components may impact the overall performance and necessitate careful consideration in the interpretation of the results.

Our experiments are also constrained by the fact that the models employed are designed for working in English only. In order to extend this approach to other languages, the multilingual variants of the NLI, claim extractor, and coreference resolution models will need to be employed. However, the performance of such multilingual models needs a proper evaluation that we defer to future work.

The claim-level granularity of the outputs of FENICE, and the alignment of claims with portions of the input document it provides, ensures a higher level of interpretability compared to previous metrics. Yet, defining a quantitative evaluation for this aspect remains an ongoing challenge.

\label{app:limitations}
\section*{Acknowledgements}
\begin{center}
\noindent
    \begin{minipage}{0.1\linewidth}
        \begin{center}
            \includegraphics[scale=0.05]{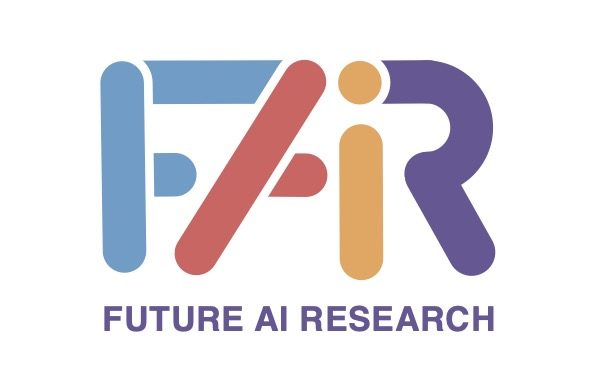}
        \end{center}
    \end{minipage}
    \hspace{0.01\linewidth}
    \begin{minipage}{0.70\linewidth}
         We gratefully acknowledge the support of the PNRR MUR project PE0000013-FAIR. 
    \end{minipage}
    \hspace{0.01\linewidth}
    \begin{minipage}{0.1\linewidth}
        \begin{center}
            \includegraphics[scale=0.08]{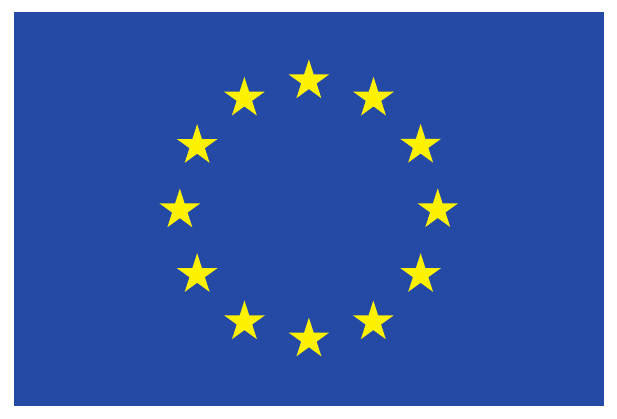}
        \end{center}
    \end{minipage}\\
\end{center}
\vspace{0.2cm}
\noindent We gratefully acknowledge the CREATIVE project (CRoss-modal understanding and gEnerATIon of Visual and tExtual content), which is funded by the MUR Progetti di Ricerca di Rilevante Interesse Nazionale programme (PRIN 2020). Alessandro Scirè and Karim Ghonim conducted this work during their enrollment in the Italian National Doctorate on Artificial Intelligence at Sapienza University of Rome.

\bibliography{main}

\cleardoublepage

\appendix




\section{Prompt for claim extraction.}
\label{app:prompt}
Figure~\ref{fig:prompt} illustrates the formulated prompt designed for extracting claims from a given summary. 
This prompt was employed to generate the claims utilized in our evaluation metric and to train our distilled claim generation model.

\section{Claim extraction evaluation with BERTScore.}
\label{app:transformer_claims}
In Table~\ref{tab:bertscore_claims}, we report the claim extraction evaluation of the LLM-based claim extractor compared to its T5-based distilled version, using BERTScore instead of ROUGE.
\begin{table}[h]
  \centering
  \small

  \begin{tabular}{cccc}
    \toprule
    Model & Precision (\%) & Recall (\%) & F1 (\%) \\
    \midrule
    GPT-3.5 & 95.8 & 94.3 & 95.0 \\
    T5$_{dist\_GPT}$ & 95.5 & 93.9 & 94.7 \\
    \bottomrule
  \end{tabular}
    \caption{Claim extractor performance comparison on ROSE~\citep{liu2023revisiting} using BERTScore~\cite{zhang2020bertscore} as backbone metric in Equations \ref{eq:easiness_p} and \ref{eq:easiness_r} instead of ROUGE.}
  \label{tab:bertscore_claims}

\end{table}

\section{NLI model ablation.}
\label{app:nli_ablation}
We evaluate the performance of \metric~utilizing RoBERTa\footnote{We load  the pre-trained model weights from the Hugging Face Hub, available at: \url{ynie/roberta-large-snli_mnli_fever_anli_R1_R2_R3-nli}.
}~\cite{liu2019roberta}
 as the NLI model for our alignments (\metric$_{RoBERTa}$), instead of DeBERTa (Table ~\ref{tab:ablation2}). 
The results reported are directly comparable to the ones presented in Table~\ref{tab:aggrefact_singlet}.

\begin{table}[h!]
    \centering
    \small
    \setlength\tabcolsep{3pt}  

    \begin{tabular}{lcccc}
        \toprule
        \textbf{Metric} & \textbf{\textsc{CNN}} & \textbf{\textsc{XSum}} & \textbf{AVG} & \textbf{$\Delta$ AVG} \\
        \midrule
        \metric & \textbf{67.5}$\pm$3.0 & \textbf{73.7}$\pm$0.3 & \textbf{70.6} & \textbf{\_} \\
        \metric$_{RoBERTa}$ & 64.4$\pm$3.2 & 68.1$\pm$1.6 & 66.3 & -4.3\\

        \bottomrule
    \end{tabular}
    \caption{Ablation study. We compare the performance of \metric~ on the \textsc{AggreFact-FtSota} split using DeBERTa and RoBERTa as the backbone NLI model.}
    \label{tab:ablation2}
\end{table}

\section{Long-form summarization factuality annotations statistics.}
\label{app:fairy_summeval_dist}
In Table~\ref{tab:fairy_summeval_dist}, we report the number of factual and unfactual summaries in our collection.
\begin{table}[H]
    \centering
    \begin{tabular}{ccl}
        \toprule
        \textbf{Factual} & \textbf{Not Factual}\\
        \midrule
        66 & 273  \\
        \bottomrule
    \end{tabular}
    \caption{Number of factual/unfactual summaries in our set of human annotations.}
\label{tab:fairy_summeval_dist}
\end{table}

\clearpage

\begin{figure}
  \begin{tcolorbox}[colback=white, colframe=black, arc=0pt, outer arc=0pt]
    \begin{varwidth}{\linewidth}
We define a claim as an "elementary information unit in a sentence, which no longer needs to be further split."\\
For example, given the following sentence:\\
INPUT:\\
NASA's Perseverance rover has discovered ancient microbial life on Mars according to a recent study published in the journal Science. It established a set of new paradigms for space exploration"\\\\ 

OUTPUT:\\
\{'claims': ["NASA's Perseverance rover discovered ancient microbial life.",\\
"Ancient microbial life was discovered on Mars.",\\
"The discovery was made according to a recent study.",\\
"The study was published in the journal Science.",\\
"The study established a set of new paradigms for space exploration."]\}\\
\\
Please consider the following recommendations:\\
If possible, use a noun as the subject in the claim (try to avoid pronouns), such as "The study established a set of new paradigms for space exploration." in the previous example.\\
Do not generate any novel word, be faithful to the provided input.\\
Your response must be directly the json, do not add any other text, such as "Here is the output:" and similar. Note that, each fact expressed in the source text must be present in the output. \\\\

Now do this task for this input\\
INPUT:    \end{varwidth}
  \end{tcolorbox}
  \caption{LLM prompt for extracting claims given a summary.}
    \label{fig:prompt}
\end{figure}

\end{document}